\documentclass[letterpaper, 10 pt, conference]{ieeeconf}  

\IEEEoverridecommandlockouts                              

\usepackage[dvipdfmx]{graphicx}
\usepackage{color}
\usepackage{float}
\usepackage{amsmath}
\usepackage{amssymb}
\usepackage{algorithm}
\usepackage{algpseudocode}
\usepackage{arydshln}
\usepackage{bm}
\usepackage{upgreek}
\usepackage{booktabs}
\usepackage{subcaption}
\usepackage{tikz}
\usepackage{cite}

\title{\LARGE \bf
Performance-Preserving Online Adaptation in Social Navigation \\ via Diffusion Steering
}

\author{Haruto Nagahisa$^{1}$$^{\dagger}$, Kohei Matsumoto$^{2}$$^{\dagger}$, Yuki Hyodo$^{1}$, and Ryo Kurazume$^{2}$
\thanks{$^{1}$Haruto Nagahisa and Yuki Hyodo are with the Graduate School of Information Science and Electrical Engineering, Kyushu University, Fukuoka, Japan {\tt\small nagahisa@irvs.ait.kyushu-u.ac.jp, hyodo@irvs.ait.kyushu-u.ac.jp}}%
\thanks{$^{2}$Kohei Matsumoto and Ryo Kurazume are with the Faculty of Information Science and Electrical Engineering, Kyushu University, Fukuoka, Japan {\tt\small matsumoto@ait.kyushu-u.ac.jp, kurazume@ait.kyushu-u.ac.jp}}%
\thanks{$^{\dagger}$Authors contributed equally}%
}

\begin{document}

\maketitle
\thispagestyle{empty}
\pagestyle{empty}

\begin{abstract}

In social navigation, modeling the complex interactions between humans and robots is difficult, and deep reinforcement learning has therefore been actively studied. However, because simulation alone cannot fully reproduce diverse scenarios, robot dynamics, and the social conventions that vary across deployment environments, fine-tuning in the deployment environment is promising. In doing so, learning that preserves the base model's performance is required, so as not to compromise the primary objective of navigation, namely avoiding pedestrians and reaching the destination. In this study, we propose a method that applies diffusion steering via reinforcement learning (DSRL), which trains only the noise policy while keeping the diffusion policy fixed, thereby achieving learning that preserves performance. Furthermore, we integrate diffusion-based RL policies trained with multiple seeds to construct the base policy, improving learning performance. Our evaluation shows that, compared with other methods, the proposed method enables efficient learning while preserving performance, and we confirm flexible behavior control through adaptation to social conventions, as well as its effectiveness on a physical robot through hardware-in-the-loop simulation.

\end{abstract}

\section{INTRODUCTION}

Social navigation, which enables navigation in crowded pedestrian environments, has emerged as a crucial task. In social navigation, appropriately modeling human-robot interactions is essential, but handcrafting these models is challenging. To address this, deep reinforcement learning (DRL) is considered an effective approach because it can implicitly learn human-robot interactions through experience. However, much of the existing work \cite{cite:SACADRL, cite:CrowdNav, cite:GCRL, cite:RGL, cite:CAWR, cite:IntensionRL, cite:COLSON} remains confined to learning in simulation, and simulation alone makes it difficult to reproduce every possible scenario or the robot's dynamics. In addition, pedestrian environments involve various social conventions that the robot is expected to follow, and these can differ depending on the deployment environment. This makes fine-tuning a pretrained model in the deployment environment a promising approach. However, realizing fine-tuning in real-world social navigation settings requires not compromising the primary objective of navigation, namely avoiding pedestrians and reaching the destination, and, to this end, requires proceeding with learning while preserving the base model's performance. In real-world environments in particular, unnecessary exploratory behavior not only delays navigation time but may also cause psychological discomfort or danger to surrounding pedestrians. \par

\begin{figure}[t]
  \centering
  \includegraphics[width=0.8\linewidth]{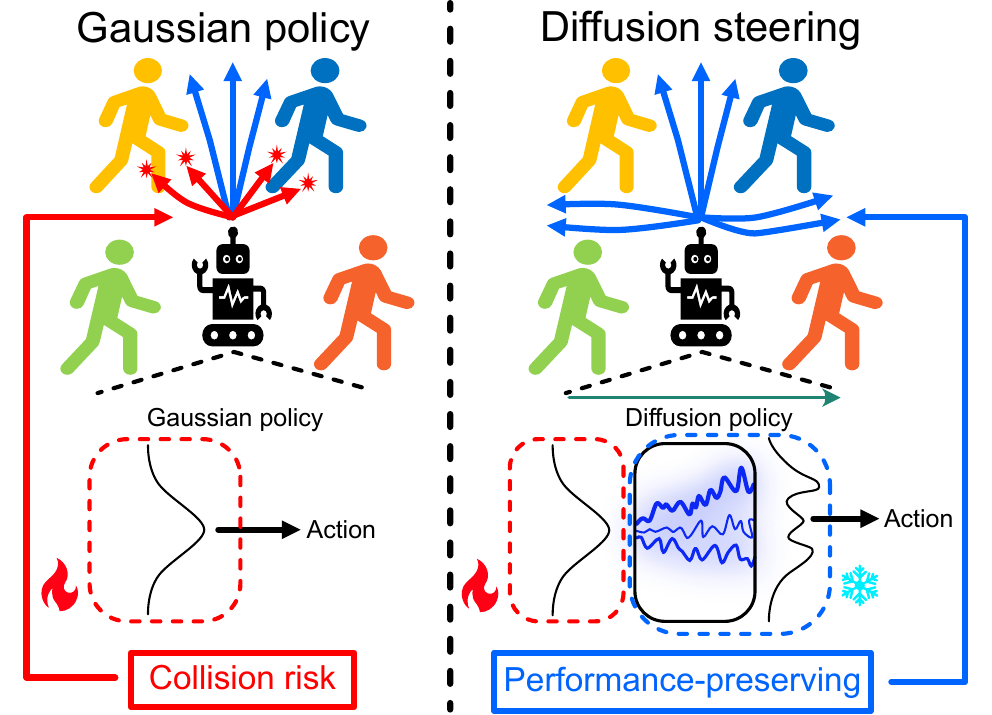}
  \caption{Conceptual illustration of the proposed method. Gaussian policies that explore the action space directly carry the risk of generating actions that collide with pedestrians during the exploration process. In contrast, diffusion steering explores the latent noise space while keeping the diffusion policy fixed, enabling learning to proceed while preserving performance.}
  \label{main_concept}
\end{figure}

Motivated by this, this study aims to learn a policy that selects the optimal action from multiple action candidates satisfying the primary objective of navigation. In RL with continuous action spaces, Gaussian distributions are commonly used for policy representation. While Gaussian distributions have the advantage of being easy to sample from, their unimodal nature prevents them from simultaneously representing multiple high-value actions. As a result, the sampled action remains merely a random perturbation of a single high-value action, and whether it satisfies the primary objective remains uncertain. In contrast, diffusion models, capable of representing multimodal action distributions, have recently achieved success as navigation policies \cite{cite:NoMaD, cite:CompoNav, cite:COLSON}. For instance, NoMaD \cite{cite:NoMaD}, which represents both goal-conditioned and exploratory actions within a single diffusion model, successfully accomplishes exploration tasks by leveraging this multimodality. Motivated by this, in this study, we focus on whether a multimodal diffusion policy can achieve learning that preserves the base model's performance, and apply diffusion steering via reinforcement learning (DSRL) \cite{cite:DSRL}, which performs RL in the latent noise space of the diffusion policy. A conceptual illustration of the proposed method is shown in Fig. \ref{main_concept}. Rather than exploring the action space directly, exploration is carried out in the latent noise space while keeping the diffusion policy fixed. This preserves performance while achieving diverse exploration through the multimodal policy distribution. Furthermore, when constructing the base diffusion policy, we propose a method that integrates multiple diffusion-based RL policies trained with different seeds. This approach achieves higher learning performance compared to methods using a single diffusion-based RL policy \cite{cite:COLSON} and methods that integrate multiple Gaussian-based RL policies \cite{cite:diffuseloco}.\par

Our main contributions are as follows:

\begin{itemize}

\item We propose a new method that applies DSRL to social navigation, with the aim of achieving learning that preserves the base model's performance. As a result, we show that, compared with other RL methods targeting diffusion policies, our proposed method is the most effective in terms of preserving the base model's performance.

\item As an application of the proposed method, we conducted training with reward designs reflecting social conventions, such as designated navigation area, keep-left rule, and keep-right rule. As a result, we confirmed that the proposed approach can adhere to these social conventions and flexibly control behavior.

\item We demonstrated that integrating multi-seed diffusion-based RL policies to construct the base diffusion policy achieves superior learning performance compared to ensembling Gaussian-based RL policies or directly employing diffusion-based RL policies.

\item Through hardware-in-the-loop (HITL) simulation experiments using a physical robot and virtual pedestrians, we demonstrated that our proposed method enables learning that preserves the base model's performance even when learning on a physical robot.

\end{itemize}

\section{RELATED WORK}

\subsection{Learning-Based Social Navigation }

In social navigation, learning-based methods have been actively studied in recent years due to the difficulty of manually modeling robot-pedestrian interactions. In particular, DRL, which learns through trial and error, has the potential to learn robot behaviors that surpass both model-based methods and those based on expert demonstration data. \cite{cite:SACADRL} is one of the earliest works to incorporate social awareness into collision avoidance. To model crowd-robot interactions, \cite{cite:CrowdNav} proposed a method that uses an attention-based pooling mechanism to learn the collective importance of neighboring pedestrians with respect to future states. In recent years, numerous methods employing Graph Neural Networks (GNNs) to learn the relationships between robots and pedestrians have also been proposed \cite{cite:RGL, cite:GCRL, cite:CAWR, cite:IntensionRL}. Furthermore, in the field of social navigation, diffusion models have begun to be utilized for pedestrian trajectory prediction and policy representation due to their strong capability in representing multimodal distributions. COLSON \cite{cite:COLSON} proposed social navigation using diffusion-based RL, demonstrating high performance and flexibility through post-hoc guidance. Other works include studies that utilized diffusion models for pedestrian trajectory prediction \cite{cite:SICNav}, and those that leveraged the compositional capabilities of diffusion models to achieve diverse action generation \cite{cite:CompoNav}. Rather than focusing on developing a new architecture for social navigation, this study builds upon existing research. Specifically, we propose a method that leverages the ability of RL to autonomously acquire high-performing policies for data generation and online learning, and integrates this with the advantage of multimodal policy representation offered by diffusion policies.

\subsection{Real-World Reinforcement Learning for Navigation}

In navigation, real-world RL is an effective strategy for addressing complex factors in real environments that are difficult to model, such as interactions with pedestrians and tire-road friction. For example, motivated by the difficulty of constructing precise vehicle dynamics simulators for high-speed navigation, \cite{cite:ResetFree} investigated real-world RL and proposed a reset-free RL approach that introduces a reset policy upon collisions or course deviations. Similarly, \cite{cite:FastRLAP} investigated vision-based navigation using a reset policy. Similar to our study, existing works have addressed the application of real-world RL to social navigation. DR-MPC \cite{cite:DRMPC} proposed a method using residual RL \cite{cite:RRL} to learn the residual actions relative to a Model Predictive Control (MPC) policy. Considering the computational resource constraints of edge devices, IRRL \cite{cite:IRRL} integrated incremental learning \cite{cite:AVG} with residual RL. Furthermore, SELFI \cite{cite:SELFI} investigated a method for learning residuals in the action-value function for vision-based social navigation. These studies focus primarily on learning in the real world alone. In addition to this, our work focuses on achieving learning that preserves the base model's performance. Specifically, we show that this can be achieved through the simple strategy of learning in the latent noise space of a fixed diffusion policy and selecting actions from the resulting multimodal diffusion policy.

\begin{figure*}[htb]
  \vspace*{2mm}
  \centering
  \includegraphics[width=0.8\textwidth]{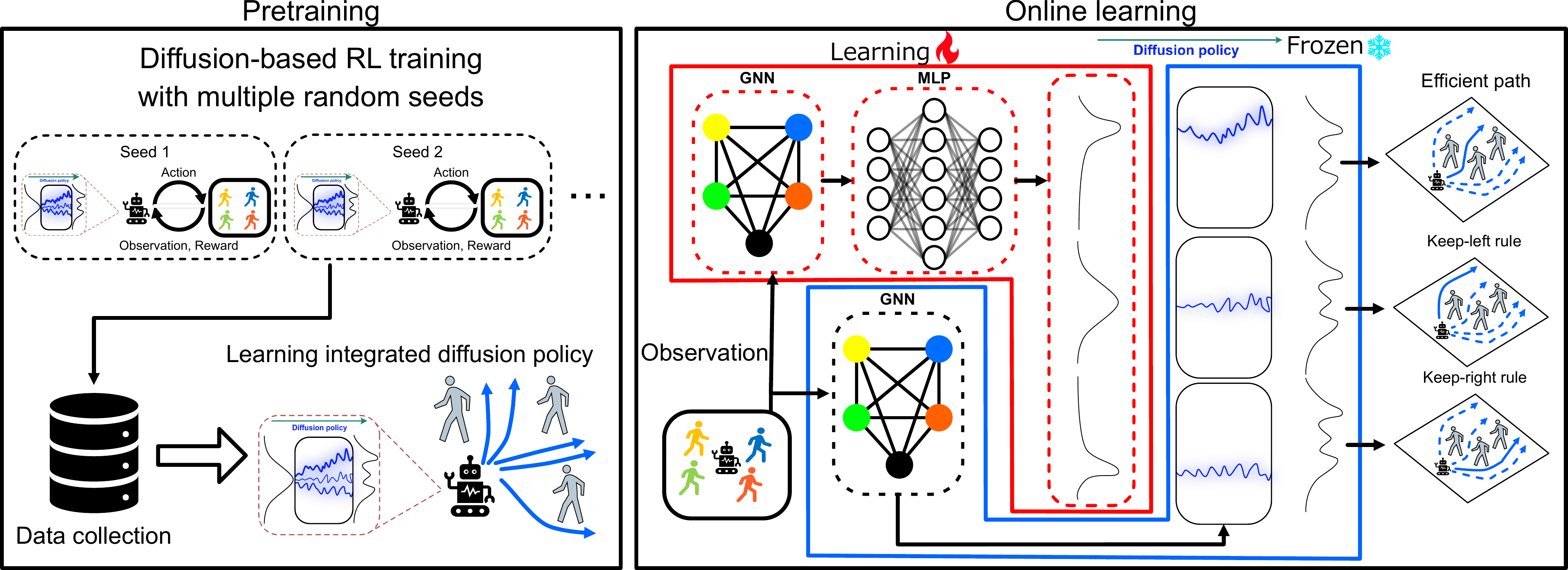}
  \caption{An overview of the proposed method. The proposed method consists of two stages: pre-training to create a base diffusion policy, and online learning using DSRL. In the pre-training stage, a diffusion policy with a multimodal policy distribution is trained by leveraging diffusion-based RL. In the online learning stage, learning is performed using DSRL while keeping the pre-trained diffusion policy fixed, enabling efficient action selection and adaptation to social conventions.
  }
  \label{approach}
\end{figure*}

\section{PRELIMINARIES}

In this study, we consider the problem of a robot aiming to reach a destination while avoiding surrounding pedestrians in an obstacle-free two-dimensional XY plane. The time step is set to 0.25 s, and the distance from the start position to the goal position is 8 m. An episode terminates if the robot collides with a pedestrian or if the episode time exceeds 30 s. At each time step t, the state of the robot is represented by the coordinates of the goal position within the robot's coordinate system, the angle to the goal position, and the robot's own velocity: $\mathbf{s}_t^r = [p_x^g, p_y^g, \theta^g, v_x, v_y]$. The state of each i-th pedestrian is represented by their position in the robot's coordinate system and their relative velocity to the robot: $\mathbf{s}_t^i = [p_x^i, p_y^i, v_x^i, v_y^i]$, and the overall state of the pedestrians is expressed as follows: $\mathbf{s}_t^h = [\mathbf{s}_t^1, \mathbf{s}_t^2,..., \mathbf{s}_t^n]$. Furthermore, since an omnidirectional mobile robot is assumed, the robot's actions are expressed as velocities in the x-direction and y-direction: $\mathbf{a}_t = [v_x, v_y]$. Finally, the reward setting in RL is defined as follows:

\begin{equation} 
r_t =
\begin{cases}
    -0.25 & \text{if } d_t^h < 0 \\
    (d_t^h-0.2)*0.125 & \text{else if } d_t^h < 0.2 \\
    1 & \text{else if } d_{t}^g \leq \rho^r \\
    0 & \text{otherwise}
\end{cases}\label{eq:reward}
\end{equation}

where $d_t^h$ is the minimum signed clearance between the robot and the pedestrians, $d_t^g$ is the distance between the robot and the goal, and $\rho^r$ is the radius of the robot. This reward structure is designed as a sparse reward, where a positive reward is granted only when the robot successfully reaches the destination while avoiding pedestrians. Furthermore, to incorporate specific social conventions into the reward design, a negative reward of $-0.25$ is applied when any of the following conditions are met. Note that left and right in the keep-left rule and keep-right rule refer to the left and right sides when viewing the environment from above.

\begin{itemize}
    \item \textbf{Designated navigation area:} The robot's position deviates beyond ±2.5 m along the lateral axis.
    \item \textbf{Keep-left rule:} The robot fails to navigate on the left side of the pedestrians.
    \item \textbf{Keep-right rule:} The robot fails to navigate on the right side of the pedestrians.
\end{itemize}

\section{APPROACH}

This section describes the learning algorithm, DSRL, as well as the pre-training and online learning phases of the proposed method. Fig. \ref{approach} shows an overview of the proposed method.

\subsection{Diffusion Steering via Reinforcement Learning}

DSRL is a method that controls the action generation of a diffusion policy by performing RL in the latent noise space of a fixed diffusion policy. Algorithm \ref{alg:dsrl_sac} shows the training procedure. In DSRL, three types of networks are trained: a policy that generates the latent noise, an action-value function that evaluates the actions generated by the diffusion policy, and a noise-value function that evaluates the latent noise. The reason for using two distinct value functions, an action-value function and a noise-value function, is as follows. In the diffusion policy, for two different noises $\mathbf{w}, \mathbf{w'} (\mathbf{w'} \neq \mathbf{w})$, it may hold that $\pi_{\mathrm{dp}}(\mathbf{s}, \mathbf{w}) \approx \pi_{\mathrm{dp}}(\mathbf{s}, \mathbf{w'})$. That is, similar actions can be generated from different latent noises, in which case directly training a value function over noise alone would be inefficient. We therefore first train the action-value function via standard TD learning, enabling it to evaluate the value of the actions generated from each noise. Next, using the output of this action-value function as the target, we train the noise-value function corresponding to the generated actions. This makes it possible to map latent noise values that have never been executed to the values of their corresponding actions that have already been executed, thereby enabling efficient learning. In this study, we use SAC \cite{cite:SAC}, an entropy-maximizing RL method, as the learning algorithm for the latent noise policy.

\begin{algorithm}
\caption{Training algorithm using DSRL with SAC}
\label{alg:dsrl_sac}
\footnotesize
\begin{algorithmic}[1]
\State \textbf{Input:} pretrained diffusion policy $\pi_{\text{dp}}$
\State Initialize replay buffer $D$, action critic $Q^A_1, Q^A_2$, latent-noise critic $Q^w_1, Q^w_2$, latent-noise actor $\pi^w$

\For{each environment step $t$}
        \State Sample latent-noise action $\mathbf{w}_t \sim \pi^w(\mathbf{s}_t)$ and 
        \Statex \hspace{1em} compute $\mathbf{a}_t \leftarrow \pi_{\text{dp}}(\mathbf{s}_t, \mathbf{w}_t)$
        \State Take action $\mathbf{a}_t$, observe $r_t$ and next state $\mathbf{s}_{t+1}$, and add $(\mathbf{s}_t, \mathbf{a}_t, r_t, \mathbf{s}_{t+1})$ to $D$
\EndFor

\For{each gradient update step}
\State Sample batch $B = {(\mathbf{s}, \mathbf{a}, r, \mathbf{s'})}$ from replay buffer $D$
\State Sample actions for computing targets $\mathbf{a'} \sim \pi_{\text{dp}}(\mathbf{s'}, \pi^w(\mathbf{s'}))$
\State Calculate targets for the action critic $Q^A$ 
\Statex \hspace{1em} $y(r, \mathbf{s'}) = r + \gamma(\mathbf{min}_{i=1, 2}Q^A_i(\mathbf{s'}, \mathbf{a'}) - \alpha \log\pi^w(\mathbf{w'}|\mathbf{s'}))$

\State Update $Q^A$: $\min \mathbb{E}\left[ (Q^A_i(\mathbf{s},\mathbf{a}) - y(r, \mathbf{s'}))^2 \right]$ for i = 1,2

\State Calculate targets for the latent-noise critic $Q^w$ 
\Statex \hspace{1em} $z(\mathbf{s}, \mathbf{w}) = \min_{i=1, 2}Q^A_i(\mathbf{s}, \pi_{\text{dp}}^w(\mathbf{s},\mathbf{w}))$

\State Update $Q^w$: 
\Statex \hspace{1em} $\min \mathbb{E}_{\mathbf{w} \sim \mathcal{N}(0,I)} \left[ (Q^w(\mathbf{s},\mathbf{w}) - z(\mathbf{s}, \mathbf{w}))^2 \right]$ for i = 1,2

\State Update $\pi^w$: $\max \mathbb{E} \left[ \min_{i=1, 2}Q^w_i(\mathbf{s}, \pi^w(\mathbf{s})) - \alpha \log\pi^w(\mathbf{w}|\mathbf{s})  \right]$

\EndFor

\end{algorithmic}
\end{algorithm}

\subsection{Diffusion Policy Using Diffusion-Based RL}

We describe the method for constructing the base diffusion policy. To enable adaptation to various environments and social conventions during training, it is desirable for the base policy to be able to represent a multimodal policy distribution. In this study, we therefore adopt an approach that integrates multiple diffusion-based RL policies trained with different seed values to construct a single diffusion policy. As the diffusion-based RL method, we adopt QSM \cite{cite:QSM}, whose effectiveness for social navigation has been demonstrated in prior work \cite{cite:COLSON}. QSM is a method that approximates the gradient of the Q-function as a score function. To train the integrated diffusion policy, we perform imitation learning using DDPM \cite{cite:DDPM} with a 100-step generation process, using data collected from the models trained with each seed. For action generation, we apply DDIM \cite{cite:DDIM} to achieve faster processing and deterministic generation, generating actions in 5 steps.

\subsection{Online Learning Using DSRL}

We describe online learning using DSRL. The objective of this study is to control the generated actions by performing RL in the latent noise space of a pretrained diffusion policy with a multimodal policy distribution, thereby achieving optimal action selection for a given environment and adaptation to social conventions. As shown in Fig. \ref{approach}, in the online learning process, only the latent noise policy is trained via RL while the weights of the pretrained diffusion policy are kept fixed. This latent noise policy uses GNNs to aggregate information about the robot and the pedestrians; specifically, we adopt GATv2 \cite{cite:GATV2}. In addition, a $\tanh$ activation function is applied to the final output layer of the latent noise policy network, followed by a scaling factor, to restrict the noise range to $[-3, 3]$. We further introduce penultimate normalization, proposed in \cite{cite:Penultimate_Normalization} as effective for stabilizing training when using $\tanh$, into both the critic and actor networks.

\section{SIMULATION EXPERIMENTS}

In this section, we verify the effectiveness of the proposed method through simulation experiments. Specifically, we evaluate the following three aspects. First, we compare our proposed method with other RL methods targeting diffusion policies to evaluate whether learning that preserves the base model's performance is achievable. Here, we use the success rate and navigation time as evaluation metrics. The success rate is an indicator of how well the base model's primary navigation objective, namely avoiding pedestrians while reaching the destination, is achieved, and it is desirable that this be maintained throughout learning. The navigation time is used to evaluate whether the robot reaches the destination without engaging in unnecessary exploratory behavior. Second, we investigate whether performance improves through training and whether behaviors adapted to social conventions have been acquired. Here, we evaluate the return from the perspective of whether learning has succeeded, and confirm the actual behavior through qualitative evaluation of trajectories. Third, focusing on the construction methods of the base diffusion policy, we compare the following three conditions: (1) a diffusion policy trained using data from multiple QSM policies with different random seeds, (2) a diffusion policy trained using data from multiple SAC policies with different random seeds, and (3) a diffusion policy trained using data from a single QSM policy. For each condition, we compare the return obtained when training with DSRL, thereby verifying the superiority in learning performance.

\begin{figure*}[htb]
  \centering
  \includegraphics[width=\textwidth]{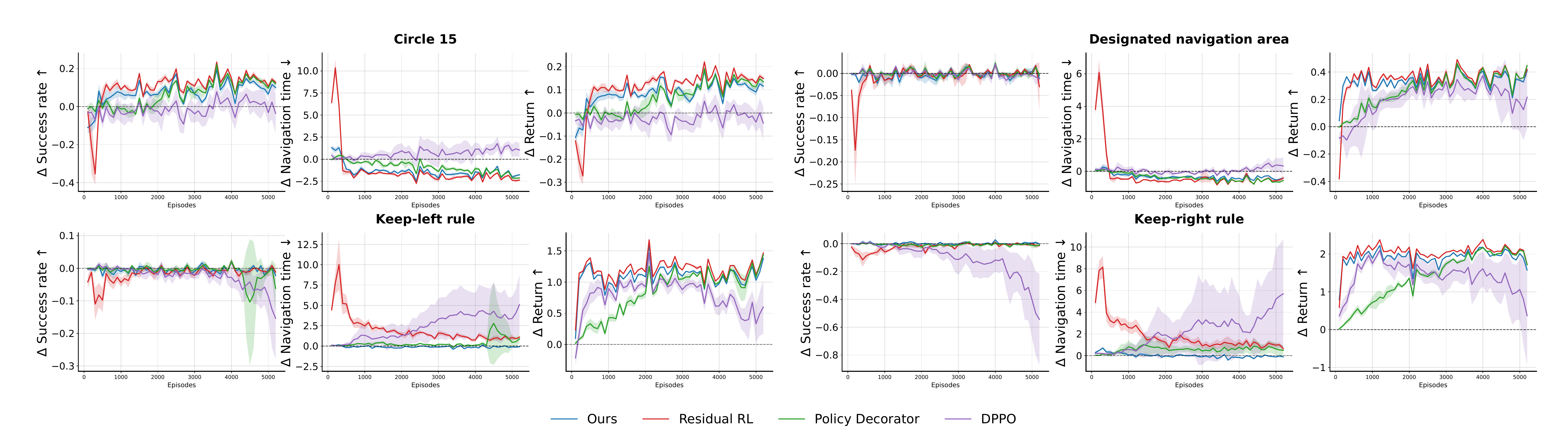}
  \caption{Learning curves showing, for each method and each scenario, the differences in success rate, navigation time, and return relative to the base diffusion policy, with the mean across seeds shown as solid lines and the standard deviation shown as shaded regions. A value of 0 on the vertical axis corresponds to the result of the base diffusion policy. Averages are plotted every 100 episodes. The corresponding scenario is indicated at the top of each graph.
  }
  \label{per_eval_sim}
\end{figure*}

\subsection{Comparative Methods}

For comparison with the proposed method, we employed RL methods utilizing Gaussian policies and methods that perform direct RL. As the methods using Gaussian policies, we adopted residual RL, which learns the residual with respect to the base policy, and Policy Decorator \cite{cite:PolicyDecorator}, which incorporates scale adjustment of the residual policy and exploration scheduling. Note that SAC was used as the underlying RL algorithm for both residual RL and Policy Decorator. For the direct RL approach, we adopted DPPO \cite{cite:DPPO}, which treats each generation step as an MDP and trains using PPO \cite{cite:PPO}. 

\subsection{Simulation Environment and Settings}

For the simulation environment in this experiment, we used CrowdNav \cite{cite:CrowdNav}, and pedestrians move according to ORCA \cite{cite:ORCA}. The maximum speed of the robot was set to 1.0 m/s. In this study, we conducted experiments from two perspectives: the validation of performance improvements and adaptation to social conventions. Training for each method was conducted for 5,000 episodes using 5 different seeds. To validate performance improvements, we used a Circle Crossing scenario populated with 15 pedestrians. This is a highly difficult environment due to the large number of pedestrians, making it a scenario where significant improvements over the base policy in terms of success rate and navigation time can be expected. Hereafter, this environment is referred to as \textit{Circle 15}. To validate adaptation to social conventions, we used a Corridor scenario populated with 5 pedestrians. In this environment, pedestrians pass through the center from both directions, providing the robot with multiple action choices, such as approaching the destination from either the left or the right, and making it suitable for evaluating flexible behavior control. In this experiment, we evaluated three conditions as social conventions: designated navigation area, keep-left rule, and keep-right rule. Hereafter, these are referred to as \textit{designated navigation area}, \textit{keep-left rule}, and \textit{keep-right rule}, respectively. \par 

For the construction of the base diffusion policy, based on preliminary experiments, we used a Square Crossing scenario populated with 30 pedestrians. In the method integrating multiple seeds, RL policies were trained using 3 different seeds, 5,000 episodes of data were collected for each seed in the same scenario, and only successful episodes were used for imitation learning. Note that in the subsequent descriptions, the methods integrating multiple seeds are referred to as \textit{multi-seed QSM} or \textit{multi-seed SAC}, depending on the base algorithm. In addition, when evaluating a single diffusion-based RL policy, we conduct evaluations using 5 different evaluation seeds for each of the 3 seed models used to construct the integrated policy, and use the results of a total of 15 trials.

\subsection{Performance Evaluation}

\textbf{1) Evaluation of Performance Preservation (Success rate and Navigation time):} First, we evaluate the success rate. As shown in Fig. \ref{per_eval_sim}, among the compared methods, residual RL and DPPO exhibit significant performance degradation, either temporarily or continuously, during training (a decrease in success rate and an increase in navigation time), failing to preserve performance. In contrast, the proposed method and Policy Decorator are able to proceed with training with almost no performance degradation in the designated navigation area and keep-right rule. On the other hand, in the remaining two scenarios, a slight performance drop is observed during the early stages of training for the proposed method in Circle 15, while Policy Decorator shows a performance drop in the keep-left rule during the later stages of training. However, considering the standard deviation, the fluctuation during these performance drops is larger for Policy Decorator, indicating that the proposed method achieves more stable training across seeds. This result shows that the proposed method narrows the exploration range of actions by exploring in the latent noise space, thereby preventing the exploration range from becoming excessively wide depending on the seed, and achieving stable learning. \par 

Next, we evaluate the navigation time. Although the proposed method shows a slight increase in navigation time during the initial stage in Circle 15, it ultimately succeeds in reducing it, and in the other scenarios it achieves a navigation time equal to or less than that of the base policy. In contrast, Policy Decorator succeeds in reducing navigation time in Circle 15 and the designated navigation area, but in the keep-left rule there is a moment where navigation time temporarily increases during the later stages of training, and in the keep-right rule navigation time continues to increase. In scenarios other than Circle 15, in addition to the sparse reward for avoiding pedestrians and reaching the destination, a reward term reflecting social conventions is added. Therefore, even if other methods show a return comparable to or higher than that of the proposed method, the strong influence of the reward term representing social conventions may lead to a solution in which the success rate and navigation time are actually worse. However, because the proposed method performs learning while keeping the diffusion policy that generates actions fixed, it is able to maintain or improve the primary objective of navigation, namely the success rate and navigation time, throughout training.\par

\begin{figure*}[htb]
  \vspace*{2mm}
  \centering
  \includegraphics[width=0.78\textwidth]{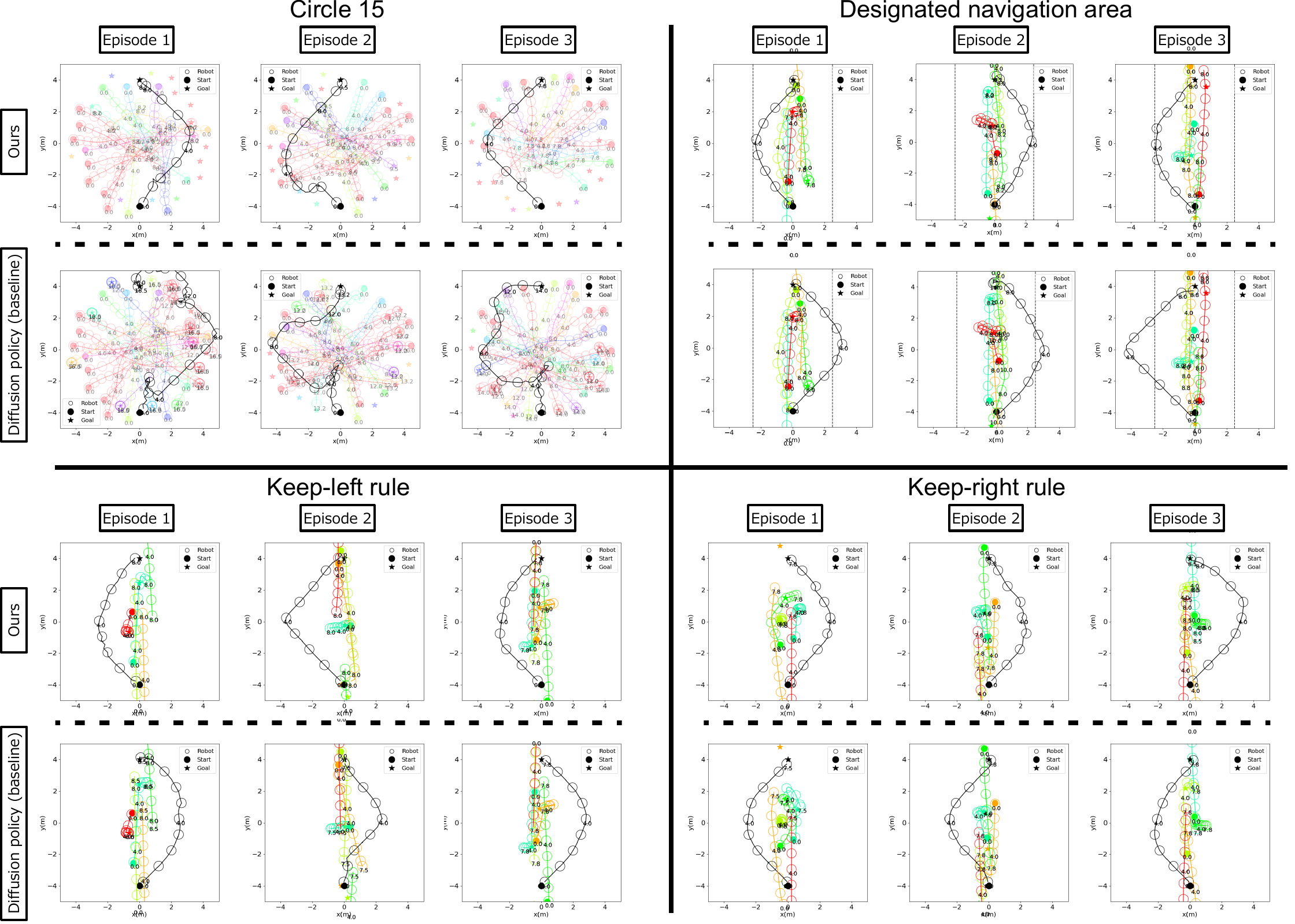}
  \caption{Trajectories for three selected episodes from the final 100 episodes of the proposed method and baseline policy in each scenario are shown. Here, the robot's trajectory is represented in black, while other colors indicate pedestrian trajectories. In scenarios with a designated navigation area, the boundary is shown with a black dotted line.
  }
  \label{per_eval_traj_sim}
\end{figure*}

\textbf{2) Evaluation of Learning Performance and Adaptation to Social Conventions (Return and Qualitative Evaluation of Trajectories):} Examining the transition of returns in Fig. \ref{per_eval_sim}, the proposed method improves the return over the base policy across all scenarios, confirming successful learning. Compared to other methods, the proposed method and residual RL achieved the fastest convergence. Although the final return is slightly higher for residual RL, this method experiences moments during training on Circle 15 where performance drops significantly and temporarily. In contrast, the proposed method succeeds in stable learning throughout all scenarios, maintaining performance without significant performance degradation. Next, through qualitative evaluation of trajectories, we examine what specific changes in behavior have resulted from training. As shown in Fig. \ref{per_eval_traj_sim}, in Circle 15, the base policy follows an inefficient path that enters the crowd before retreating, whereas the proposed method is able to follow an efficient path that detours around the crowd from the initial stage. Furthermore, across the three scenarios evaluating adaptation to social conventions (designated navigation area, keep-left rule, and keep-right rule), the proposed method appropriately adheres to conventions that were ignored by the base policy.

\begin{figure}[htb]
  \centering
  \includegraphics[width=0.5\textwidth]{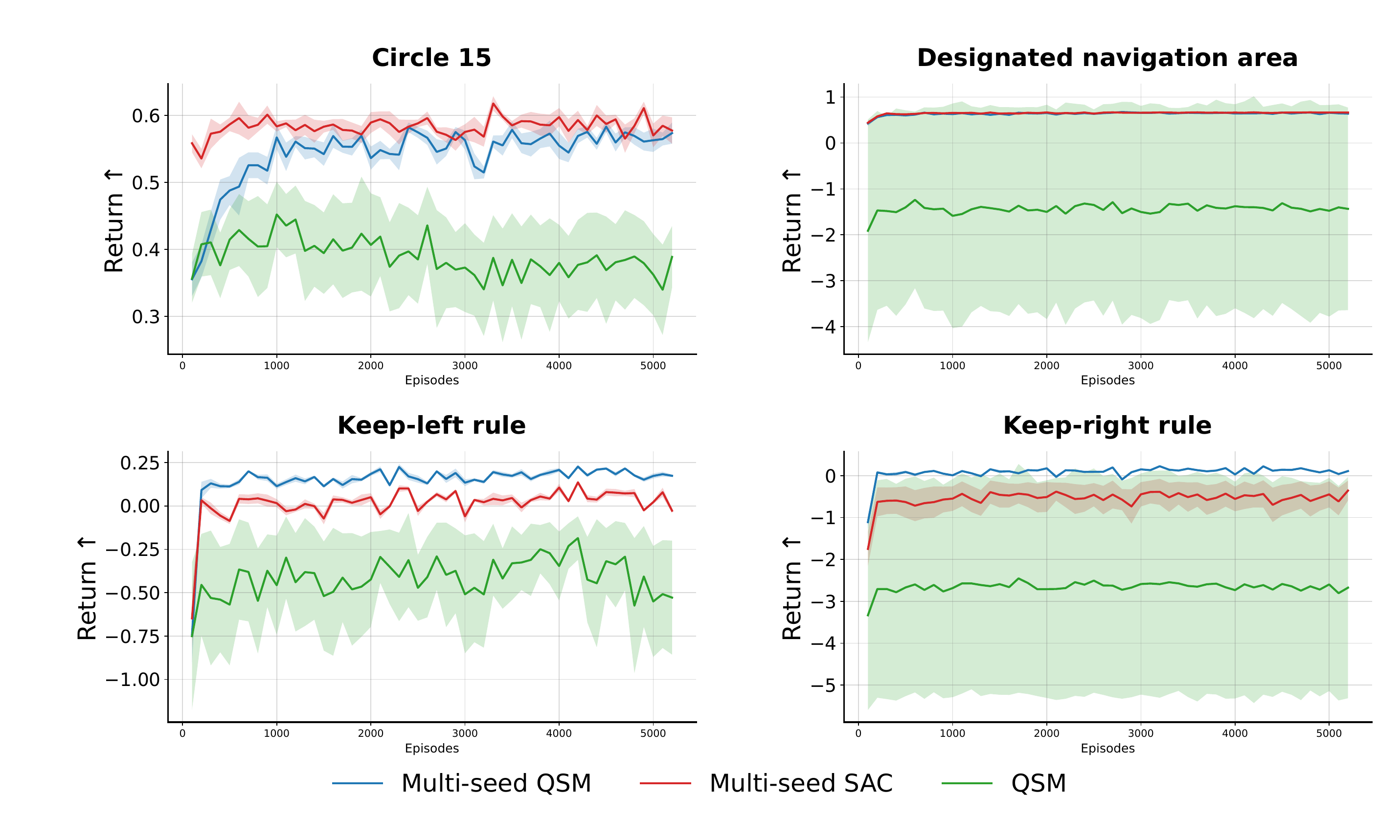}
  \caption{Learning curves showing, for each scenario, the return trained using DSRL for the three methods used to construct the base diffusion policy, with the mean across seeds shown as solid lines and the standard deviation shown as shaded regions. Values are plotted as averages over every 100 episodes. Scenarios are indicated at the top of each plot.
  }
  \label{base_compare}
\end{figure}

\textbf{3) Comparison of Base Policy Construction Methods (Return):} Finally, we compare the construction methods of the base diffusion policy. As shown in Fig. \ref{base_compare}, multi-seed QSM achieves higher final performance and smaller standard deviation than single QSM across all scenarios, demonstrating its superior learning performance and stability. This results is attributable to the fact that integrating models from multiple seeds suppresses the bias inherent to each individual model while incorporating a more diverse range of behaviors, thereby achieving high multimodality. Furthermore, in comparison with multi-seed SAC, although multi-seed SAC marginally outperforms in Circle 15, multi-seed QSM significantly surpasses it in both the keep-left rule and the keep-right rule. This results shows that using diffusion-based RL policies achieves higher multimodality than Gaussian policies, thereby enabling a wider range of action selection.

\section{HARDWARE-IN-THE-LOOP EXPERIMENTS}

\begin{figure}[htb]
  \vspace*{2mm}
  \centering
  \includegraphics[width=0.3\textwidth]{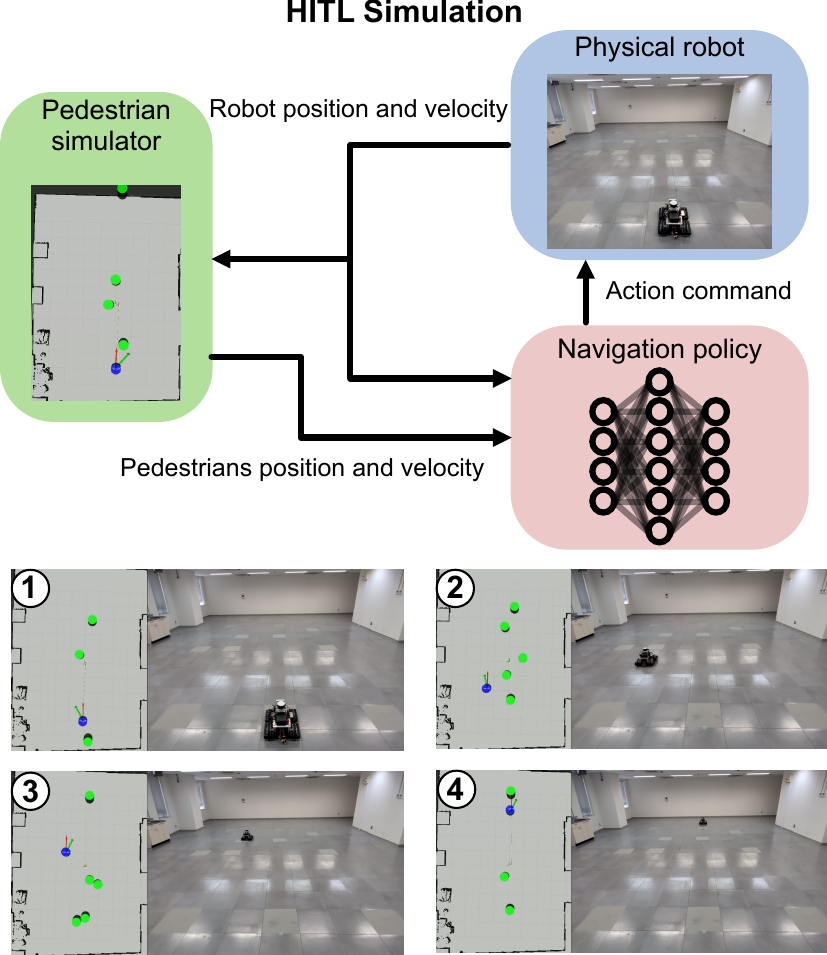}
  \caption{The upper figure provides an overview of the HITL simulation system, while the lower figure illustrates its operation. In the lower figure, the left side displays the RViz2 visualization, where green cylinders represent pedestrians and the blue cylinder represents the robot. The right side shows the robot operating in the physical environment.
  }
  \label{hitl_simulation}
\end{figure}

In this section, we verify the proposed method through hardware-in-the-loop (HITL) simulation using a physical robot and virtual pedestrians. Because it is difficult from a resource perspective to constantly prepare multiple pedestrians and repeat the training process in a real-world environment, we adopted a HITL simulation as the experimental environment to verify the learning on a physical robot. An overview of the framework and its operation are shown in Fig. \ref{hitl_simulation}. Based on the physical robot's specifications, the maximum velocity was set to 0.5 m/s, and experiments were conducted within a constrained space measuring 6 m × 9 m long. In this experiment, we evaluate whether learning that preserves the base model's performance and flexible behavior control during learning are achievable. For this purpose, we set up two environments requiring the keep-left rule and the keep-right rule, respectively, within a Corridor scenario featuring 5 pedestrians. Training for each method was conducted over 500 episodes; upon completing each episode, model updates were performed for a number of times equal to the number of steps collected in that episode. As the comparison method, we adopted Policy Decorator, which achieved the second-best results after our proposed method in terms of performance-preserving learning in the simulation experiments. We compare and evaluate the transitions of success rate, navigation time, and return during the learning processes of both methods.

\subsection{Hardware and Software Design}

This system is based on a rover equipped with mecanum wheels. An onboard computer (Jetson AGX Orin) handles sensor data acquisition from a 3D-LiDAR (AR-XT-32-a) used for measurement, as well as processes such as self-localization. Meanwhile, inference and training are executed on a remote computer equipped with an Intel Core Ultra 9 285K CPU and an NVIDIA GeForce RTX 5080 GPU. 

\subsection{Performance Evaluation}

\begin{figure}[htb]
  \centering
  \includegraphics[width=0.5\textwidth]{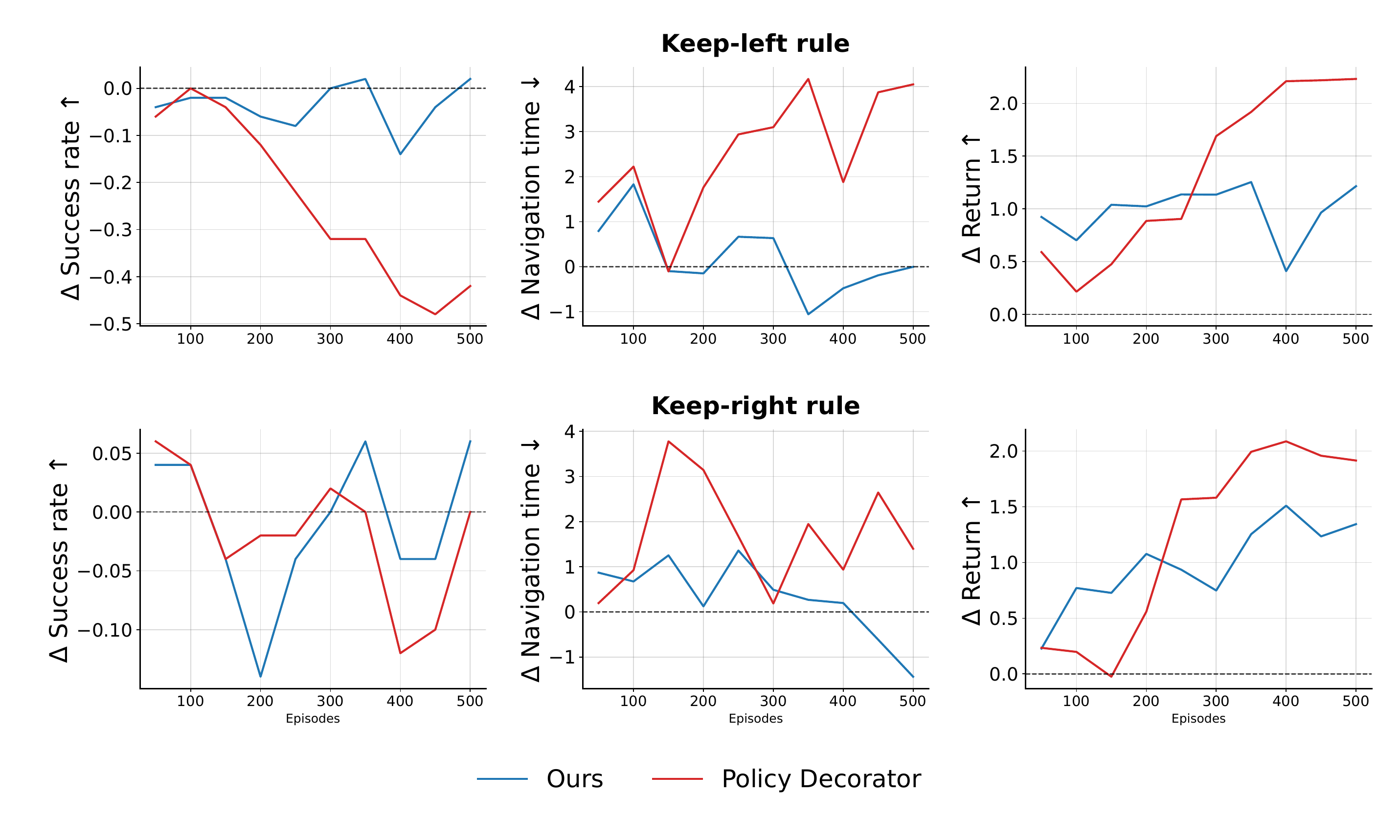}
  \caption{Learning curves showing the differences in success rate, navigation time, and return between each method and the baseline diffusion policy across scenarios. A value of 0 on the vertical axis corresponds to the result of the base diffusion policy. Values are averaged over every 50 episodes, with the scenario indicated at the top of the plot.
  }
  \label{real_compare}
\end{figure}

As shown by the return curves in Fig. \ref{real_compare}, it can be confirmed that the proposed method consistently improves performance compared to the base policy. Next, comparing the proposed method with Policy Decorator, the proposed method proceeds with training while minimizing drops in success rate and increases in navigation time in both the keep-left rule and the keep-right rule. In contrast, Policy Decorator suffers a substantial drop in success rate in the keep-left rule and exhibits worse navigation time across both scenarios. On the other hand, while the proposed method and Policy Decorator demonstrated comparable performance in terms of return in the previous simulation experiments, Policy Decorator achieved higher performance in this experiment. This can be attributed to the fact that, due to the specifications of the physical robot, the maximum velocity in this experiment was restricted to a lower value than in the simulation environment. This restriction likely degraded the multimodality of the diffusion policy, consequently reducing the number of available action candidates. However, as with the results of the simulation experiments, the proposed method is able to achieve learning that maintains the success rate and navigation time, and it can be said that it shows excellent results with respect to the objective of this study.\par

\begin{figure}[htb]
  \vspace*{2mm}
  \centering
  \includegraphics[width=0.3\textwidth]{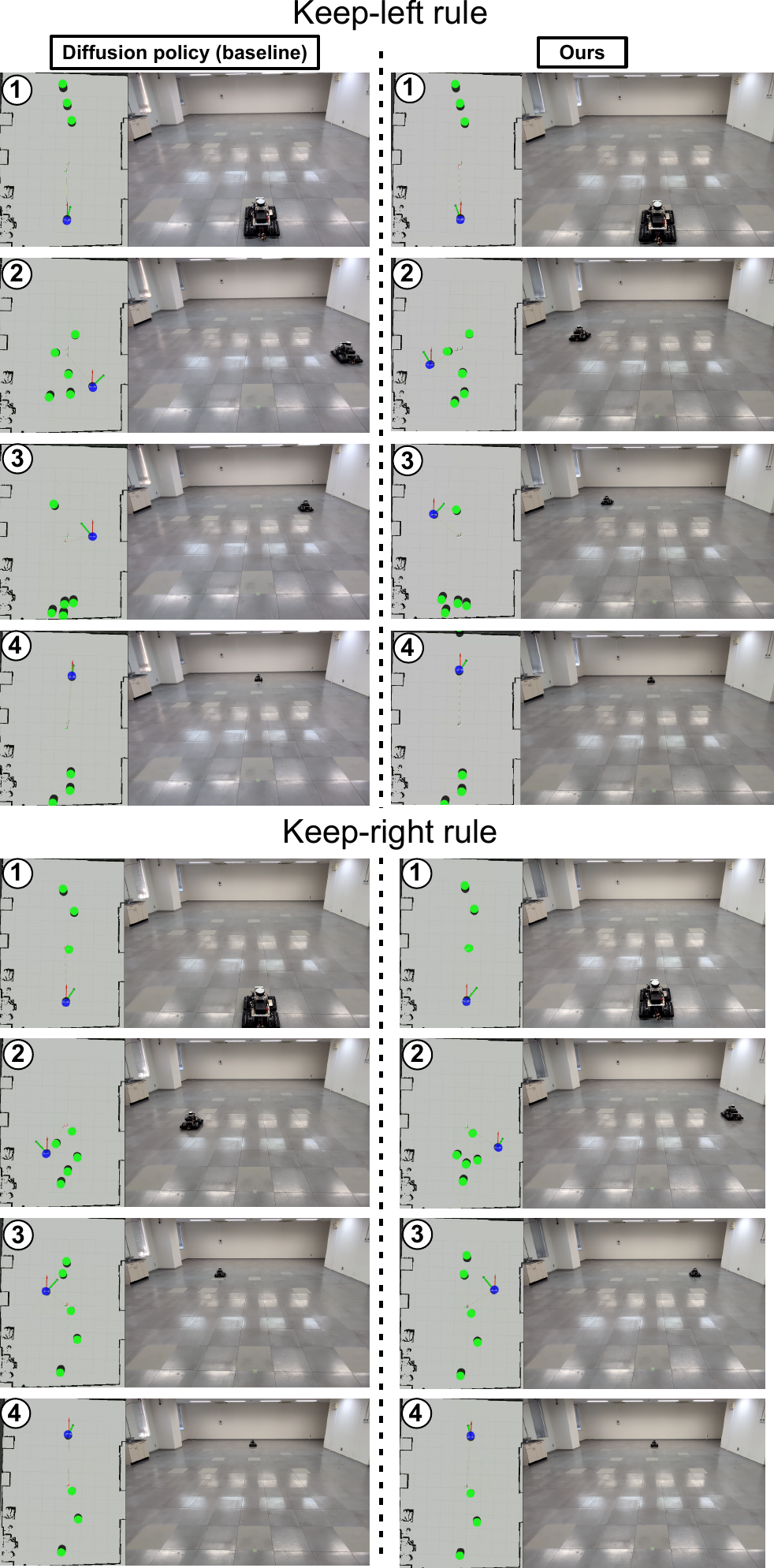}
  \caption{Comparison of the behavior of the proposed method and the base diffusion policy in one episode selected from the final 100 episodes, for each scenario.
  }
  \label{real_traj_compare}
\end{figure}

Furthermore, as shown in Fig. \ref{real_traj_compare}, in both scenarios the proposed method appropriately adapts, through training, to the social conventions that were ignored by the base policy prior to learning, achieving navigation that complies with these rules.

\section{CONCLUSIONS}

In this study, we proposed a method that applies DSRL to a multimodal diffusion policy, aiming to achieve learning that preserves the base model's performance in social navigation. Through our evaluation, we demonstrated that the proposed method achieves learning that preserves performance compared with other methods, that, as an application of the proposed method, the robot's behavior can be adapted to social conventions, and that integrating multiple diffusion-based RL policies improves learning performance. In addition, results from the HITL simulation demonstrated that our method is able to preserve the base model's performance even when learning on a physical robot. \par

However, this study has not yet validated the learning process in environments with actual pedestrians. Because real-world deployments introduce challenges such as difficulties in pedestrian recognition and self-localization in dynamic environments, verifying the proposed method in actual pedestrian environments remains a subject for future work.

\bibliographystyle{IEEEtran}
\bibliography{IEEEabrv,reference}

\end{document}